\documentclass[letterpaper, 10 pt, conference]{ieeeconf}  

\IEEEoverridecommandlockouts                              

\usepackage{graphics} 
\usepackage{epsfig} 
\usepackage{mathptmx} 
\usepackage{times} 
\usepackage{amsmath} 
\usepackage{amssymb}  
\usepackage{booktabs} 
\usepackage{multirow}  
\usepackage{makecell}  
\usepackage{graphicx}
\usepackage{graphicx}
\usepackage[utf8]{inputenc}
\usepackage{amsmath}
\usepackage{amssymb}
\usepackage{booktabs} 
\usepackage{multirow} 
\usepackage{url}
\usepackage{xcolor}
\usepackage{algorithm}
\usepackage[noend]{algpseudocode} 
\usepackage{hyperref} 
\usepackage{cite}     
\usepackage{lipsum}   

\title{\LARGE \bf
CAMF: Collaborative Adversarial Multi-agent Framework for Machine Generated Text Detection}

\author{Yue Wang$^{\text{1}}$, Liesheng Wei $^{\text{2}}$ and Yuxiang Wang$^{\text{3}*}$\thanks{$^*$Corresponding author.} 
\thanks{$^{\text{1}}$Yue Wang is with Stanford University, Stanford, CA, USA. (Email: {\tt\small wyue0125@alumni.stanford.edu})}%
\thanks{$^{\text{2}}$Liesheng Wei is with the College of Information Technology, Shanghai Ocean University, Shanghai, China. (Email: {\tt\small alvinwei1024@gmail.com})}%
\thanks{$^{\text{3}}$Yuxiang Wang is with the School of Business, Stevens Institute of Technology, Hoboken, NJ, USA. (Email: {\tt\small yuxiang.wang0476@gmail.com})}
}

\begin{document}

\maketitle
\thispagestyle{empty}
\pagestyle{empty}

\begin{abstract}
Detecting machine-generated text (MGT) from contemporary Large Language Models (LLMs) is increasingly crucial amid risks like disinformation and threats to academic integrity. Existing zero-shot detection paradigms, despite their practicality, often exhibit significant deficiencies. Key challenges include: (1) superficial analyses focused on limited textual attributes, and (2) a lack of investigation into consistency across linguistic dimensions such as style, semantics, and logic. To address these challenges, we introduce the \textbf{C}ollaborative \textbf{A}dversarial \textbf{M}ulti-agent \textbf{F}ramework (\textbf{CAMF}), a novel architecture using multiple LLM-based agents. CAMF employs specialized agents in a synergistic three-phase process: \emph{Multi-dimensional Linguistic Feature Extraction}, \emph{Adversarial Consistency Probing}, and \emph{Synthesized Judgment Aggregation}. This structured collaborative-adversarial process enables a deep analysis of subtle, cross-dimensional textual incongruities indicative of non-human origin. Empirical evaluations demonstrate CAMF's significant superiority over state-of-the-art zero-shot MGT detection techniques.
\end{abstract}

\section{Introduction}

Large language models (LLMs), enabled by rapid deep learning advancements\cite{li2025graph,li2025graph,li2025privacy,tong2024mmdfnd,lu2025dammfnd,liu2025coherency,he2022novel,li2024mhhcr,liu2025mdn,cui2025diffusion,mo2024min,mo2025one,cui2025multi}, have shown exceptional ability in generating human-like text, transforming numerous applications. However, their widespread use introduces significant risks, including enabling cyber threats, spreading misinformation \cite{weidinger2021ethical}, undermining academic honesty \cite{perrin2015academic}, and introducing code vulnerabilities \cite{pearce2021natural}. Furthermore, the statistical artifacts of MGT risk contaminating future training datasets, a phenomenon known as model poisoning \cite{shumailov2023curse}. Consequently, developing robust methodologies for detecting machine-generated text (MGT) is crucial for mitigating these potential harms.

The task of MGT detection is typically a binary classification problem: distinguishing human-authored text from LLM-generated text \cite{solan2023detecting}. Current detection strategies include \emph{feature-based detection}, which relies on internal model states (e.g., logits) \cite{gehrmann2019gltr} that are often inaccessible, necessitating imprecise surrogates \cite{guo2023close}. \emph{Supervised fine-tuning} trains classifiers on labeled data but suffers from poor generalization and overfitting to unseen models or domains \cite{solan2023detecting,jaiswal2023detecting}. \emph{Zero-shot textual analysis}, which examines intrinsic linguistic properties without model-specific training \cite{mitchell2023detectgpt,jaiswal2023zero}, is favored for its practicality. However, these methods have critical limitations.

Existing zero-shot detection techniques often exhibit notable weaknesses:
\begin{enumerate}
    \item \textbf{Superficial Textual Analysis:} Many methods focus on surface-level features \cite{wang2023prompted} or semantic continuity \cite{zhou2024ai}. This narrow focus fails to capture the complex interplay of textual characteristics, limiting their effectiveness against nuanced MGT, such as partially generated or human-edited text.
    \item \textbf{Absence of Cross-Dimensional Consistency Verification:} Current techniques often lack rigorous verification of consistency across linguistic dimensions. Superficially coherent text may hide subtle syntactic, lexical, or logical deviations that indicate an artificial origin.
\end{enumerate}

To address these limitations, we propose the \textbf{C}ollaborative \textbf{A}dversarial \textbf{M}ulti-agent \textbf{F}ramework (\textbf{CAMF}). CAMF operates on the hypothesis that MGT, despite its fluency, contains subtle incongruities across stylistic, semantic, and logical dimensions that are exposed through collaborative and adversarial scrutiny. Inspired by multi-agent paradigms for reasoning and debate \cite{du2023improving,guan2025mmd}, CAMF decomposes detection using specialized LLM-based agents in a structured three-stage process: \emph{Multi-dimensional Linguistic Feature Extraction}, \emph{Adversarial Consistency Probing}, and \emph{Synthesized Judgment Aggregation}. Initially, agents profile the text's linguistic features. Then, they adversarially probe these profiles for inconsistencies indicative of MGT. Finally, a synthesis agent integrates the findings for a final judgment.

Our primary contributions are summarized as follows:
\begin{itemize}
    \item We introduce \textbf{CAMF}, a novel multi-agent framework that applies a structured collaborative and adversarial system to MGT detection.
    \item We design a three-stage process (\emph{Multi-dimensional Linguistic Feature Extraction}, \emph{Adversarial Consistency Probing}, \emph{Synthesized Judgment Aggregation}) enabling systematic, multi-faceted analysis and in-depth consistency verification.
    \item We empirically demonstrate CAMF's superiority over state-of-the-art zero-shot MGT detectors and validate its components via rigorous ablation studies.
\end{itemize}

\section{Related Work}
\label{sec:related_work}

\subsection{Machine-Generated Text Detection}
The growth of fluent LLMs requires effective machine-generated text (MGT) detection to counter risks like disinformation and dataset contamination (\cite{weidinger2021ethical,shumailov2023curse}). While active methods like watermarking (\cite{kirchenbauer2023watermark}) have deployment hurdles, passive detection of output text is vital. Passive methods are either white-box, requiring often-infeasible internal model access (\cite{gehrmann2019gltr}), or black-box, using only the text. Within black-box methods, zero-shot techniques (\cite{mitchell2023detectgpt,jaiswal2023zero}) are preferred for their generalizability, yet they often rely on superficial features (\cite{wang2023prompted}) and lack adequate assessment of cross-dimensional linguistic consistency (\cite{zhou2024ai}). These shortcomings hinder the detection of subtle MGT. Our CAMF framework addresses these deficiencies with a multi-agent architecture for in-depth analysis and consistency verification.

\subsection{Multi-Agent Systems}
Multi-agent systems (MAS) using LLMs have become a powerful paradigm, often outperforming single agents\cite{li2025comprehensive}. They tackle complex problems via structured collaboration (\cite{hong2023metagpt}) and enhance reasoning through adversarial processes like debate and critique (\cite{chan2023chateval,du2023improving}). These systems use LLM reasoning for task decomposition and solution synthesis. Inspired by the success of collaborative-adversarial frameworks in boosting performance (\cite{guan2025mmd,park2024predict}), CAMF applies these multi-agent principles to MGT detection. To our knowledge, CAMF is the first application of a structured collaborative-adversarial MAS, using specialized agents, for the specific challenge of discriminating between human and machine-generated text.

\section{Methodology}
\label{sec:methodology}

This section provides a detailed exposition of the \textbf{C}ollaborative \textbf{A}dversarial \textbf{M}ulti-agent \textbf{F}ramework (\textbf{CAMF}), our proposed architecture for discriminating machine-generated text (MGT) from human-authored content. CAMF is architected to overcome the limitations of existing zero-shot detectors, namely their superficial textual analysis and lack of cross-dimensional consistency verification. Drawing inspiration from multi-agent collaboration and adversarial reasoning \cite{hong2023metagpt,du2023improving,park2024predict}, CAMF orchestrates specialized Large Language Model (LLM)-based agents across three synergistic stages. This structured approach facilitates a nuanced, multi-faceted textual examination to uncover subtle incongruities indicative of artificial generation.

\begin{figure*}[htbp] 
    \centering 
    \includegraphics[width=0.9\textwidth]{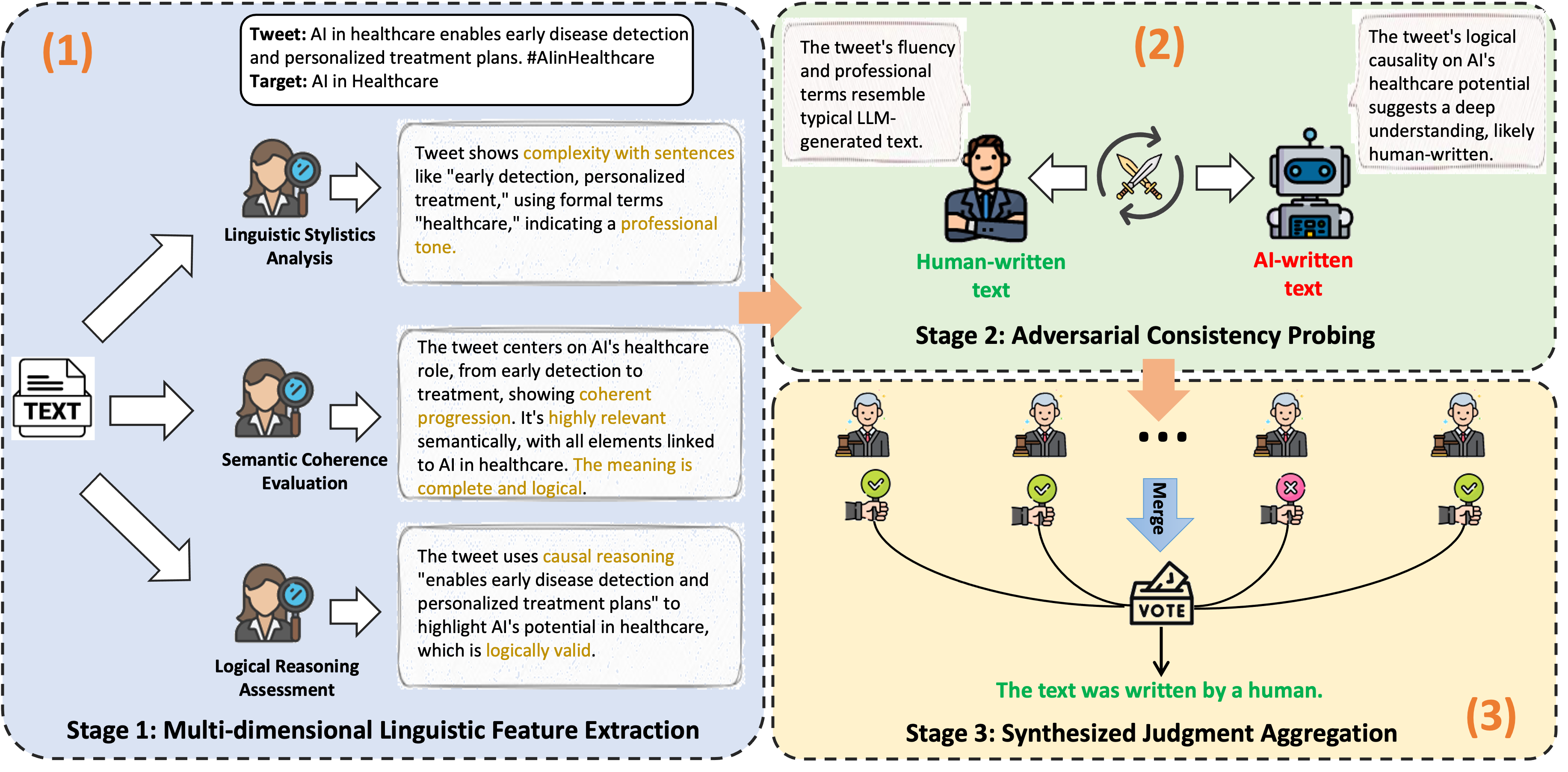} 
    \caption{Architectural Overview of the CAMF Framework for Machine-Generated Text Detection. } 
    \label{fig:framework} 
\end{figure*}

\subsection{Problem Definition}

Consistent with established MGT detection research \cite{solan2023detecting}, we define the task as a binary classification problem. Given an input text sequence $x$, the objective is to predict its authorship label $y \in \{0, 1\}$, where $y=0$ signifies human authorship and $y=1$ indicates LLM generation. The goal is to develop a robust detector function, $f: x \mapsto \hat{y}$, that accurately generalizes to unobserved text instances across diverse models and domains.

\subsection{Framework Overview}
The CAMF architecture, depicted in Figure \ref{fig:framework}, systematically decomposes the MGT detection task into three interconnected stages:

\begin{enumerate}
    \item \textbf{Multi-dimensional Linguistic Feature Extraction:} This initial stage employs specialized agents to analyze the input text $x$ from distinct linguistic perspectives. They generate comprehensive analytical profiles, capturing stylistic, semantic, and logical traits, directly addressing the need for analysis beyond superficial features.
    \item \textbf{Adversarial Consistency Probing:} Building on the initial profiles, this stage introduces an adversarial dynamic. Agents rigorously probe for consistency across linguistic dimensions, aiming to uncover subtle incongruities that might betray an MGT origin.
    \item \textbf{Synthesized Judgment Aggregation:} In the final stage, a dedicated synthesis agent integrates the multi-dimensional profiles and the refined insights from the adversarial probing to render a final, consolidated classification decision.
\end{enumerate}

\subsection{Stage 1: Multi-dimensional Linguistic Feature Extraction}
\label{subsec:stage1_extraction}

This foundational stage confronts the limitation of superficial textual analysis by constructing a rich, multi-faceted representation of the input text $x$. It leverages specialized agents to avoid reliance on a narrow set of indicators, enabling the detection of more subtle MGT signals. We instantiate three profiling agents:

\subsubsection{Linguistic Stylistics Analysis ($\mathcal{A}_{LS}$)}
Agent $\mathcal{A}_{LS}$ acts as a computational stylistics expert. It meticulously analyzes the text $x$ for features of writing style, syntax, and lexical usage. It scrutinizes aspects like syntactic complexity (e.g., sentence length distributions, parse tree depth), lexical diversity (e.g., Type-Token Ratio), and characteristic usage of linguistic markers (e.g., function words, modal verbs) to discern patterns differentiating MGT from human writing. Its output is a comprehensive textual profile, $P_{LS}(x)$, summarizing these stylistic observations.
\begin{equation}
    P_{LS}(x) \leftarrow \text{Analyze}_{\text{Style}}(x; \theta_{LS})
    \label{eq:stylistic_analysis}
\end{equation}
where $\text{Analyze}_{\text{Style}}$ is the LLM-driven stylistic analysis process guided by parameters $\theta_{LS}$.

\subsubsection{Semantic Coherence Evaluation ($\mathcal{A}_{SC}$)}
As a semantic analyst, agent $\mathcal{A}_{SC}$ evaluates the continuity and logical progression of meaning in text $x$. Its purpose is to identify semantic irregularities like abrupt topic shifts, subtle contradictions, excessive redundancy, or an unnaturally predictable thematic flow often seen in MGT. It assesses thematic development and the logical integrity of information. The outcome is a detailed textual profile, $P_{SC}(x)$, evaluating the text's semantic coherence.
\begin{equation}
    P_{SC}(x) \leftarrow \text{Evaluate}_{\text{Coherence}}(x; \theta_{SC})
    \label{eq:semantic_analysis}
\end{equation}
where $\text{Evaluate}_{\text{Coherence}}$ represents the agent's synthesis process focused on semantic consistency under parameters $\theta_{SC}$.

\subsubsection{Logical Reasoning Assessment ($\mathcal{A}_{RL}$)}
Agent $\mathcal{A}_{RL}$ evaluates the logical structure and argumentative rigor within the input text $x$. It examines the validity of arguments, the linkage between evidence and claims, and aims to detect logical fallacies or overly simplistic or conspicuously flawless logical structures sometimes observed in LLM outputs. The resultant textual profile, $P_{RL}(x)$, encapsulates this comprehensive logical assessment.
\begin{equation}
    P_{RL}(x) \leftarrow \text{Assess}_{\text{Logic}}(x; \theta_{RL})
    \label{eq:logical_analysis}
\end{equation}
where $\text{Assess}_{\text{Logic}}$ signifies the LLM-guided evaluation of logical soundness, using criteria defined by $\theta_{RL}$.

Collectively, these agents generate a set of analytical profiles $\{P_{LS}(x), P_{SC}(x), P_{RL}(x)\}$, constituting a multi-dimensional 'fingerprint' of the text $x$ that provides the foundation for the subsequent stage.

\subsection{Stage 2: Adversarial Consistency Probing}
\label{subsec:adversarial_probing}

This stage directly addresses the lack of rigorous cross-dimensional consistency verification in many MGT detectors. It operates on the premise that authorship signals often manifest in the subtle harmony—or dissonance—across linguistic dimensions. Let $P = \{P_{LS}(x), P_{SC}(x), P_{RL}(x)\}$ be the profiles from Stage 1. This stage uses a structured adversarial interaction between two specialized agents to actively probe for inconsistencies, enhancing robustness against sophisticated MGT.

\subsubsection{Adversarial Argument Generation ($\mathcal{A}_{GM}$)}
Agent $\mathcal{A}_{GM}$, the Generator-Mimic, receives the profile set $P$. It identifies potential weaknesses or contradictions across the profiles ($P_{LS}, P_{SC}, P_{RL}$) to construct a counter-argument. Its objective is to generate an adversarial textual argument, $R_{GM}(x)$, that highlights these potential inconsistencies, simulating a "best effort" challenge to the initial assessment.
\begin{equation}
    R_{GM}(x) \leftarrow \text{GenerateArgument}(P; \theta_{GM})
    \label{eq:adversarial_argument}
\end{equation}
where $\text{GenerateArgument}$ symbolizes the LLM-driven process focused on articulating cross-profile inconsistencies within $P$.

\subsubsection{Consistency Refinement ($\mathcal{A}_{DE}$)}
Agent $\mathcal{A}_{DE}$, the Detector-Enhancer, acts as a critical evaluator. It receives both the original profiles $P$ and the adversarial argument $R_{GM}(x)$. Its core task is to rigorously evaluate the validity of the challenges in $R_{GM}(x)$ by cross-referencing them against the full evidence in $P$. $\mathcal{A}_{DE}$ resolves apparent conflicts and produces a refined, adversarially-informed analysis report, $R_{DE}(x)$.
\begin{equation}
    R_{DE}(x) \leftarrow \text{RefineAnalysis}(P, R_{GM}(x); \theta_{DE})
    \label{eq:consistency_refinement}
\end{equation}
where $\text{RefineAnalysis}$ denotes the LLM-guided process of reconciling the initial profiles with the adversarial challenge.

The output of this stage, the pair $(R_{GM}(x), R_{DE}(x))$, encapsulates both the constructed challenge and the system's refined assessment, serving as crucial input for the final judgment in Stage 3.

\subsection{Stage 3: Synthesized Judgment Aggregation}
\label{subsec:stage3_aggregation}

The culminating stage of CAMF consolidates the rich and potentially conflicting information into a single, definitive classification. This crucial function is performed by the \textbf{Synthesis Judge Agent ($\mathcal{A}_{SJ}$)}, acting as an impartial adjudicator.

The $\mathcal{A}_{SJ}$ receives the complete set of inputs: the multi-dimensional profiles from Stage 1, $\{P_{LS}(x), P_{SC}(x), P_{RL}(x)\}$, and the adversarial outputs from Stage 2, $R_{GM}(x)$ and $R_{DE}(x)$. Its core responsibility is to meticulously weigh the strength and consistency of all presented evidence. It evaluates indicators supporting human authorship ($y=0$) against those favoring machine generation ($y=1$), critically considering how effectively the detector-enhancer ($\mathcal{A}_{DE}$) addressed the challenges posed by the generator-mimic ($\mathcal{A}_{GM}$). The agent synthesizes this information to identify which authorship hypothesis provides a more compelling and parsimonious explanation. Based on this holistic synthesis, $\mathcal{A}_{SJ}$ renders the final prediction $\hat{y}$. This decision-making process is formalized as:
\begin{equation}
    \hat{y} = f_{SJ}(P_{LS}(x), P_{SC}(x), P_{RL}(x), R_{GM}(x), R_{DE}(x)) \in \{0, 1\}
    \label{eq:final_judgment}
\end{equation}
where $f_{SJ}$ represents the complex reasoning logic implemented by the LLM guiding the $\mathcal{A}_{SJ}$. Through this structured, three-stage process, CAMF provides a more robust, interpretable, and accurate methodology for MGT detection compared to approaches reliant on monolithic analysis.

\section{Experiments}
\label{sec:experiments}

\subsection{Experiments Setting}

\subsubsection{Datasets}
We use five diverse datasets from the \textit{Raidar} benchmark \cite{mao2024raidar}: \texttt{News}, \texttt{Creative Writing}, \texttt{Student Essay}, \texttt{Code}, \texttt{Reviews}. Each dataset comprises paired human-authored texts and MGT counterparts (primarily from GPT-3.5-turbo, GPT-4), spanning distinct domains. These datasets facilitate focused analyses for RQ2-RQ4, while all five are used for the comprehensive performance evaluation (RQ1).

\subsubsection{Baseline Methods}
We benchmark \textbf{CAMF} against a representative selection of state-of-the-art zero-shot MGT detectors. This includes: commercial systems like \textbf{GPTZero} \cite{gptzero}; probability-based techniques such as \textbf{DetectGPT} \cite{mitchell2023detectgpt} and \textbf{Ghostbuster} \cite{verma2023ghostbuster}; a perturbation-based approach, \textbf{Raidar} \cite{mao2024raidar}; and competitive LLM-based baselines using \texttt{gpt-4o}: direct prompting (\textbf{GPT4o}), chain-of-thought (\textbf{GPT+CoT}), and a ReAct-inspired multi-agent setup (\textbf{GPT+React}) \cite{yao2023react}.

\subsubsection{Evaluation Metrics}
The primary metric for detection performance is the macro \textbf{F1-Score}, chosen for its balanced consideration of precision and recall. We also report \textbf{Accuracy} to provide a comprehensive performance picture.

\subsubsection{Implementation Details}
All agents within CAMF ($\mathcal{A}_{LS}$, $\mathcal{A}_{SC}$, $\mathcal{A}_{RL}$, $\mathcal{A}_{GM}$, $\mathcal{A}_{DE}$, $\mathcal{A}_{SJ}$) were implemented using \texttt{gpt-3.5-turbo} via its API. For reproducibility, the sampling \texttt{temperature} and \texttt{top\_p} parameters were set to 0. The agents' roles and interactions strictly adhere to the three-stage process. The number of adversarial probing rounds was set to 2 by default. Baselines were implemented based on their official descriptions or public implementations for equitable comparison.

\subsection{Overall Performance}

To address RQ1, we evaluated CAMF against baselines across the full suite of five datasets. Table \ref{tab:overall_performance} summarizes the performance results.

\begin{table*}[htbp]
\centering
\caption{Overall Performance Comparison (\textbf{F1-Score} / \textbf{Accuracy}). CAMF consistently demonstrates superior performance compared to baseline methods across diverse datasets. }
\label{tab:overall_performance}
\renewcommand{\arraystretch}{1.1}
\setlength{\tabcolsep}{5pt}
\begin{tabular}{lccccc}
\toprule
\multirow{2}{*}{\textbf{Methods}} & \textbf{News} & \textbf{Creative Writing} & \textbf{Student Essay} & \textbf{Code} & \textbf{Reviews}  \\
& (F1 / Acc) & (F1 / Acc) & (F1 / Acc) & (F1 / Acc) & (F1 / Acc) \\
\midrule
GPTZero        & 62.15 / 63.41 & 65.22 / 66.18 & 63.58 / 64.82 & 88.76 / 89.03 & 81.29 / 82.11 \\
DetectGPT      & 61.52 / 62.78 & 64.98 / 65.91 & 62.33 / 63.59 & 87.91 / 88.24 & 80.55 / 81.46 \\
Ghostbuster    & 60.88 / 62.13 & 63.75 / 64.77 & 61.89 / 63.17 & 86.54 / 86.98 & 79.12 / 80.09 \\
Raidar         & 63.18 / 64.39 & 66.55 / 67.42 & 64.21 / 65.38 & 90.11 / 90.46 & 82.56 / 83.33 \\
GPT4o          & 70.53 / 71.66 & 74.11 / 74.98 & 72.87 / 73.79 & 93.05 / 93.21 & 86.81 / 87.48 \\
GPT+CoT        & 71.89 / 72.95 & 75.46 / 76.29 & 74.22 / 75.11 & 93.64 / 93.88 & 87.44 / 88.06 \\
GPT+React      & 72.52 / 73.58 & 76.09 / 76.91 & 74.85 / 75.73 & 94.27 / 94.51 & 88.07 / 88.68 \\
\textbf{CAMF (Ours)} & \textbf{74.67} / \textbf{75.59} & \textbf{78.31} / \textbf{79.08} & \textbf{76.94} / \textbf{77.81} & \textbf{95.18} / \textbf{95.37} & \textbf{90.43} / \textbf{90.96} \\
\bottomrule
\end{tabular}
\end{table*}

The empirical results in Table \ref{tab:overall_performance} unequivocally demonstrate CAMF's superior detection capability, as it consistently achieves the highest \textbf{F1-Score} and \textbf{Accuracy} across all datasets. Significantly, CAMF surpasses even the capable \texttt{gpt-4o} powered baselines. On average, CAMF achieves an \textbf{F1-Score} improvement of 2.15 percentage points over the strongest baseline (GPT+React) and 12.41 points over Raidar. This outcome strongly underscores the effectiveness of CAMF's structured multi-agent architecture, which enables more nuanced detection than direct prompting or alternative multi-agent designs.

\subsection{Ablation Study}

To evaluate the contribution of individual components within CAMF (RQ2), we conducted a systematic ablation study. We assessed the following variations against the complete model on the \texttt{News}, \texttt{Student Essay}, and \texttt{Code} datasets:
\begin{itemize}
    \item \textbf{w/o LS}, \textbf{w/o SC}, \textbf{w/o RL}: CAMF excluding the Linguistic Stylistics, Semantic Coherence, or Logical Reasoning agent, respectively.
    \item \textbf{w/o Adversarial Probing}: CAMF removing the entire Stage 2. The Synthesis Judge decides based directly on Stage 1 profiles.
    \item \textbf{w/o Synthesis Judge}: CAMF removing the final Stage 3. Classification is determined heuristically from prior stage outputs.
\end{itemize}

\begin{figure}[htbp]
    \centering
    \includegraphics[width=1.0\columnwidth]{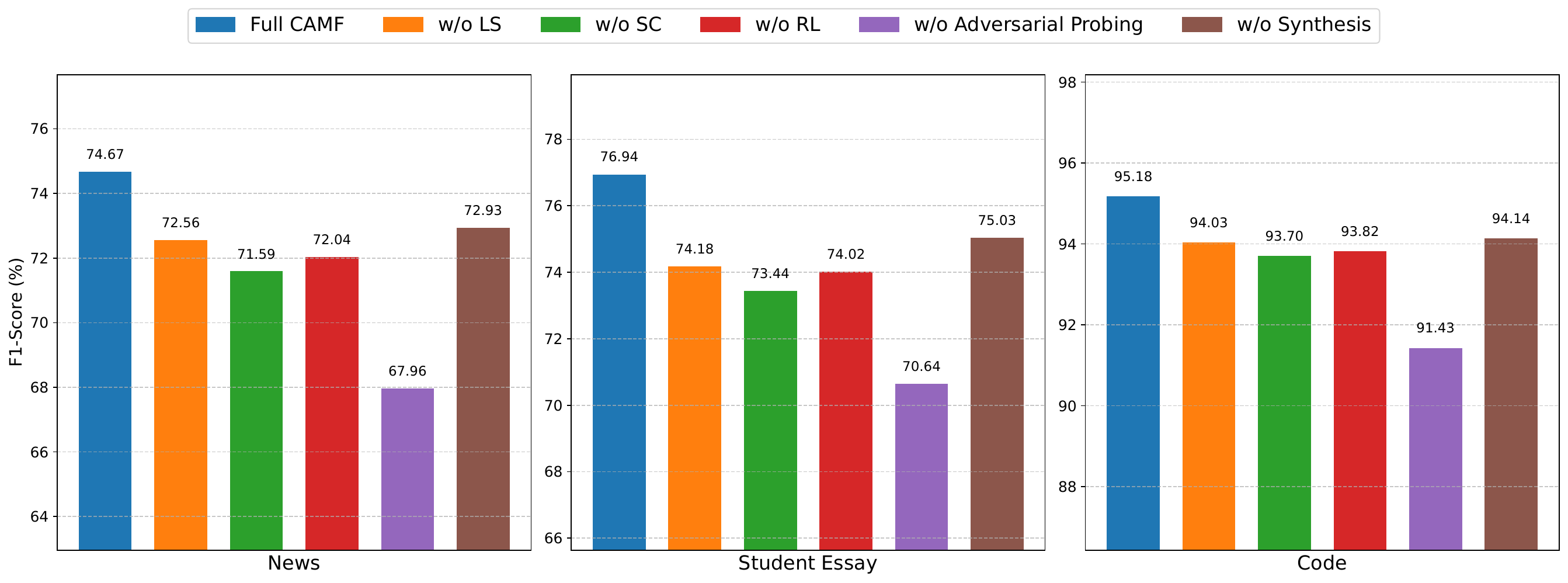}
    \caption{Ablation study results.}
    \label{fig:ablation}  
\end{figure}

As illustrated in Figure \ref{fig:ablation}, removing any single component leads to a discernible decrease in \textbf{F1-Score} across all tested datasets, validating that each part contributes synergistically to CAMF's overall efficacy. Notably, the most substantial performance degradation occurs when the \emph{Adversarial Consistency Probing} stage is ablated, underscoring its critical importance for uncovering subtle, cross-dimensional inconsistencies. Removing individual profiling agents also results in significant performance drops, indicating that each linguistic dimension provides non-redundant information. While removing the \emph{Synthesized Judgment Aggregation} stage yields the smallest relative decrease, the degradation remains consistent, highlighting the benefit of a dedicated mediator agent. These findings affirm the effective synergistic design of CAMF.

\subsection{Impact of Adversarial Probing Rounds (RQ3)}

\begin{figure}[htbp]
    \centering
    \includegraphics[width=1.0\columnwidth]{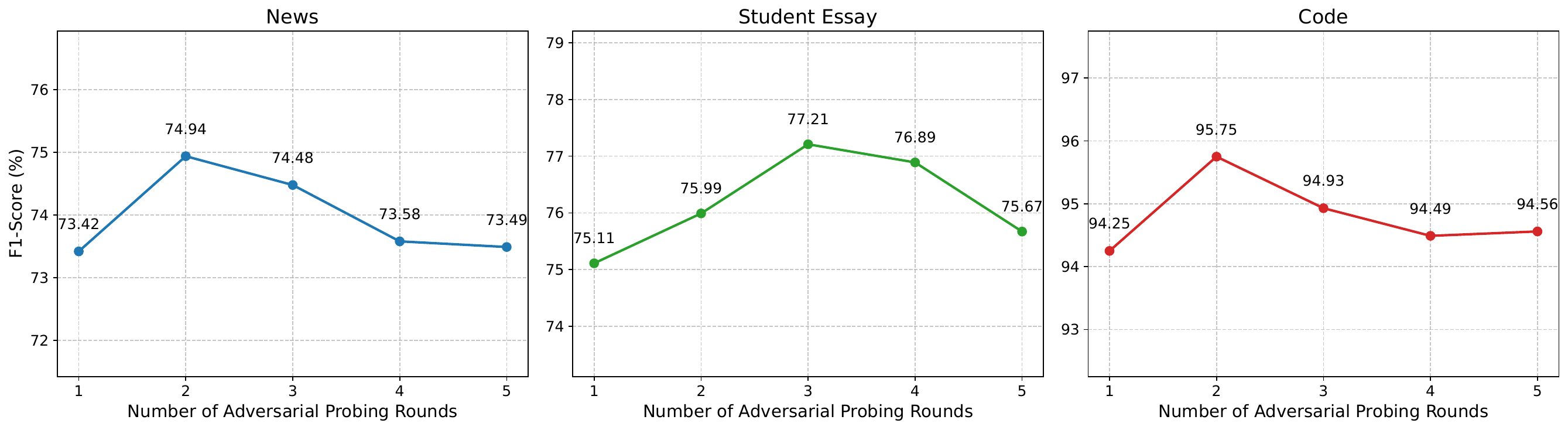}
    \caption{Impact of the Number of Adversarial Probing Rounds on \textbf{F1-Score} (\%). Performance generally peaks around 2-3 rounds, suggesting an optimal balance of analytical thoroughness and efficiency for these datasets.}
    \label{fig:debate_rounds}  
\end{figure}

To investigate the influence of adversarial interaction depth (RQ3), we varied the number of interaction rounds in Stage 2. Each round is an exchange cycle between the Generator-Mimic ($\mathcal{A}_{GM}$) and Detector-Enhancer ($\mathcal{A}_{DE}$) agents. We measured the \textbf{F1-Score} for 1 through 5 rounds on three representative datasets, with results in Figure \ref{fig:debate_rounds}.

Figure \ref{fig:debate_rounds} indicates that increasing probing rounds from a single round generally enhances performance, likely due to more thorough adversarial analysis. However, the benefits appear to saturate or slightly diminish beyond 2-3 rounds for these datasets. Performance on \texttt{News} and \texttt{Code} peaks at 2 rounds, while \texttt{Student Essay} peaks at 3. This pattern suggests an optimal trade-off: sufficient rounds are necessary, but an excessive number might introduce diminishing returns or noise. Based on these findings, we adopted 2 rounds as the default configuration, representing a good balance between performance and efficiency.

\subsection{Impact of Backbone LLM Selection (RQ4)}
\label{subsec:backbone_llm_choice}

\begin{figure}[htbp]
    \centering
    \includegraphics[width=1.0\columnwidth]{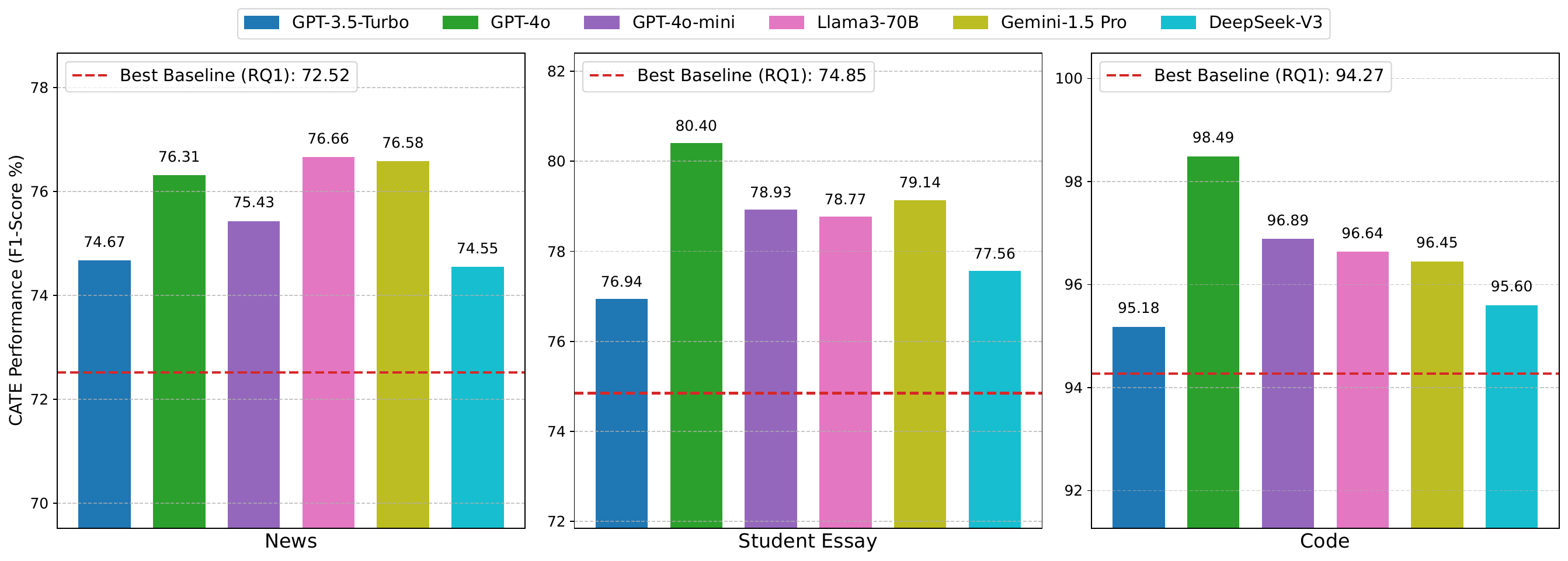}
    \caption{Impact of Backbone LLM Choice on CAMF's Detection Performance (\textbf{F1-Score} \%). While absolute performance scales with the underlying LLM, all CAMF variants significantly outperform the best baseline methods from Table \ref{tab:overall_performance}, underscoring the architecture's robustness.}
    \label{fig:backbone_llm_impact}  
\end{figure}

To evaluate CAMF's reliance on the underlying LLM engine (RQ4), we assessed its performance with different backbones: \texttt{GPT-3.5-Turbo} (our default), the more advanced \texttt{GPT-4o}, and \texttt{Llama3-70B}. The objective was to understand how performance scales with LLM capability and whether CAMF's architectural benefits persist. Performance results (\textbf{F1-Score}) are depicted in Figure \ref{fig:backbone_llm_impact}.

The results in Figure \ref{fig:backbone_llm_impact} show that, as expected, CAMF's absolute \textbf{F1-Score} improves with more powerful foundational LLMs like \texttt{GPT-4o}. However, the most significant finding is that the CAMF framework consistently and substantially outperforms all evaluated baselines (Table \ref{tab:overall_performance}) regardless of the backbone LLM used. This demonstrates the inherent architectural strength and robustness of CAMF's collaborative adversarial approach. The gains are not solely dependent on using the most advanced LLM; the structured interaction itself provides significant advantages. Our default choice of \texttt{GPT-3.5-Turbo} confirms CAMF can deliver state-of-the-art results with more accessible and cost-effective models.

\subsection{Inference Time Analysis}
\label{subsec:inference_time}

We evaluated inference efficiency alongside detection performance on the \texttt{News} dataset (Table~\ref{tab:inference_time}). CAMF achieves the highest accuracy with a competitive inference time (22.5s). It demonstrates a clear advantage over \textbf{GPT+React} \cite{yao2023react}, being significantly faster while more accurate. CAMF is also faster than \textbf{Raidar} \cite{mao2024raidar} (25.1s) and only slightly slower than the much less accurate \textbf{DetectGPT} \cite{mitchell2023detectgpt} (19.5s). Notably, direct inference using large models like \texttt{o1} (45.0s) and \texttt{deepseek-r1} (64.7s) was substantially slower and did not reach CAMF's performance levels. These results highlight CAMF's effective balance between state-of-the-art accuracy and practical inference speed.

\begin{table}[htbp]
\centering
\caption{Comparison of Average Per-Sample Inference Time (seconds) and Performance (\textbf{F1-Score} / \textbf{Accuracy} \%) on the \texttt{News} dataset. CAMF demonstrates the best performance with competitive speed.}
\label{tab:inference_time}
\renewcommand{\arraystretch}{1.1} 
\setlength{\tabcolsep}{8pt} 
\begin{tabular}{lcc}
\toprule
\textbf{Method} & \textbf{Avg. Inference Time (s)} & \textbf{F1 / Acc (\%)} \\
\midrule
DetectGPT \cite{mitchell2023detectgpt} & 19.5 & 61.52 / 62.78 \\
Raidar \cite{mao2024raidar} & 25.1 & 63.18 / 64.39 \\
GPT+React \cite{yao2023react}  & 31.8 & 72.52 / 73.58 \\
\textbf{CAMF} & \textbf{22.5} & \textbf{74.67 / 75.59} \\
\midrule 
\texttt{o1}  & 45.0 & 71.0 / 70.2 \\ 
\texttt{deepseek-r1}  & 64.7 & 72.4 / 72.3 \\ 
\bottomrule
\end{tabular}
\end{table}

\section{Conclusion}
\label{sec:conclusion}

To overcome the superficiality and lack of cross-dimensional analysis in current machine-generated text (MGT) detection, we introduced the \textbf{C}ollaborative \textbf{A}dversarial \textbf{M}ulti-agent \textbf{F}ramework (\textbf{CAMF}). By utilizing specialized LLM agents for \emph{Multi-dimensional Linguistic Feature Extraction}, \emph{Adversarial Consistency Probing}, and \emph{Synthesized Judgment Aggregation}, CAMF enables deep scrutiny of subtle textual incongruities and enhances robustness through adversarial refinement. Empirical evaluations rigorously demonstrated CAMF's superiority over state-of-the-art zero-shot methods, and ablation studies validated the contribution of its synergistic, adversarial design. This work highlights the efficacy of structured multi-agent systems for the complex MGT detection challenge. Future directions include handling mixed human-AI content, exploring broader domain adaptability, and enhancing detection through more meticulous reasoning.


\end{document}